# Why Trust in AI May Be Inevitable

Nghi Truong

Sasin School of Management, Chulalongkorn University, nghi.truong@sasin.edu

Phanish Puranam

INSEAD, phanish.puranam@insead.edu

Ilia Tsetlin

INSEAD, ilia.tsetlin@insead.edu

In human-AI interactions, explanation is widely seen as necessary for enabling trust in AI systems. We argue that trust, however, may be a pre-requisite because explanation is sometimes impossible. We derive this result from a formalization of explanation as a search process through knowledge networks, where explainers must find paths between shared concepts and the concept to be explained, within finite time. Our model reveals that explanation can fail even under theoretically ideal conditions - when actors are rational, honest, motivated, can communicate perfectly, and possess overlapping knowledge. This is because successful explanation requires not just the existence of shared knowledge but also finding the connection path within time constraints, and it can therefore be rational to cease attempts at explanation before the shared knowledge is discovered. This result has important implications for human-AI interaction: as AI systems, particularly Large Language Models, become more sophisticated and able to generate superficially compelling but spurious explanations, humans may default to trust rather than demand genuine explanations. This creates risks of both misplaced trust and imperfect knowledge integration.

## 1 INTRODUCTION

As Artificial Intelligence (AI) systems based on deep neural networks take on more consequential decision-making roles, their lack of transparency has led to ethical concerns and user mistrust. Explanation - providing justification (i.e. reasons) for accepting a proposition - is increasingly seen as useful for building trust in artificial intelligence systems [22], [17]. A common prescription is that by providing clear explanations of how AI systems work and why specific decisions were made, developers can increase understanding and foster trust among users [25]. This perspective implicitly assumes that explanation precedes and enables trust - we trust what we understand.

While we do not question the importance of explanation for trust-building, it is also true that we must sometimes trust because something cannot be explained. Using a formal model, we show that explanation can fail even when actors are rational, honest, motivated, can communicate perfectly, and possess overlapping knowledge. As a consequence, trust in another's competence [14] or benevolence [13], can become critical for one actor to accept and use information from another. In this perspective, trust is a substitute for explanation.

The intuition for our result is as follows: we know that successful explanation requires knowledge overlap between parties [8], [27] – but we point out that this overlap must first be discovered. We formalize this idea by modeling explanation as a search process, where successful explanation requires finding paths in "knowledge graphs" - networks of knowledge-elements - between what the explainer knows in common with the explainee, and what needs to be explained. This search is inherently challenging because the explainer must implicitly navigate two distinct knowledge structures – their own and their estimate of the explainee's – within finite time. This gives rise to our result on the limits to explanation even under theoretically ideal conditions.

The limits to explanation are particularly relevant to human-AI interactions, where AI agents represent and process knowledge differently from human agents, making the search for common knowledge even more challenging. This result, we argue, has two important implications. First, when facing these explanation limits, humans may default to trust as an alternative mechanism. However, while trust can enable knowledge utilization in the short term, it is an inferior substitute for explanation because it does not enable the human to fully integrate new knowledge with existing understanding. Without such integration, the recipient cannot independently apply, adapt, or effectively transmit the knowledge [24]. Second, using trust as a substitute for explanation is particularly dangerous in settings where there is a risk of human's misplaced trust in AI arising from spurious explanations. This risk has intensified with developments in Large Language Models-LLMs [38]. By gaining trust through facile explanations today, an LLM may lead humans to reduce their demand for explanations from it in the future.

The rest of this paper is organized as follows. We first review prior related work on explanation in both computer and organizational science. We then describe the model and derive the key results in two succeeding sections. A final section discusses the implications for human-AI interaction.

## 2 RELATED LITERATURE

### 2.1 Explanation as a source of trust

Prior literature has identified several distinct mechanisms through which explanation may foster trust. First, explanations help users understand AI systems as intentional agents with beliefs and goals, bridging the gap between users' folk understanding of AI behavior and its actual operation [22]. Second, explanations contribute to establishing "common ground" – shared understanding between human and AI actors that enables effective interaction [17]. Third, explanation serves as a mechanism for transparency, allowing users to evaluate the system's decision-making logic [25].

The literature has also identified several properties of effective explanations in human-AI interaction. Explanations should be contrastive, addressing specific user queries about why one decision was made over alternatives [22]. They should be interactive, enabling dialogue and follow-up questions rather than one-way information transfer [20]. Perhaps most importantly, explanations must be calibrated to users' prior knowledge and expectations – what is obvious to an AI expert may be incomprehensible to a domain specialist in another field [7].

This body of work thus shares the assumption that explanation precedes and enables trust – that we trust what we understand. While this assumption is intuitively appealing, we argue it may be incomplete: trust may also be required because explanation is difficult. To understand the challenge of explanation,

## 2.2 Barriers to explanation

Explanation plays an important role in the transfer and integration of knowledge (justified true belief – [24]). Since knowledge often resides within individuals, knowledge creation and deployment in organizations often relies on transferring knowledge across individuals [11], [40].

Prior research has identified four major barriers to such knowledge transfer. First, the explainer and explainee may have misaligned incentives, leading to strategic withholding of information or insufficient effort in knowledge sharing [4]. Second, explainers may possess tacit knowledge that resists articulation – they may know more than they can tell [26], [23], [35]. Third, even when knowledge is explicit and parties are motivated to share, the transfer process can be impeded by communication challenges, including both basic channel noise that distorts message transmission [29] and deeper semantic differences in how parties interpret shared information [3]. Fourth, the explainee may lack the related knowledge necessary to absorb and integrate new information, making transfer impossible even when knowledge is clearly conveyed [8], [28], [27]. These barriers can operate independently or in combination, creating persistent challenges for knowledge integration in organizations.

While remedies exist for each of these challenges, several authors recognize that there are others for which there may be no remedy- making explanation forever unattainable [9]. In fact this limit to explainability justifies reliance on authority (e.g., [[34], [10], [16]) and trust [14], [13]. However, this literature does not offer a clear account for why such irreducible barriers to explanation might exist among honest and motivated actors. Given finite cognitive capacities, no individual is omniscient; yet why they are heterogeneous remains to be explained rather than asserted, particularly if there are strong incentives for both parties to become homogenous in what they know, knowledge is explicit, and communication channels are noise-free. If bounded rationality manifested itself primarily as a capacity constraint, preventing any individual from being omniscient, then actors with initially different knowledge might nonetheless – given the right incentives and communication channels – converge on the same knowledge. If they choose not to do so, their incentives are, by definition, misaligned.

We believe an important reason for the persistent heterogeneity of even explicit knowledge among boundedly rational, but honest and motivated actors equipped with perfect communication channels is that the absorption and assimilation of new information typically requires the ability to form connections with what is already known (a point well recognized in prior literature – e.g., [8]), and that discovering these points of connection within finite time may be impossible (a point which is not).

For instance, in his theory of explanation, Thagard [36], [37] emphasizes that explanations coherent with recipients' prior beliefs are more likely to be accepted. In an influential paper, Cohen and Levinthal [8] also highlight the importance of existing related knowledge for absorbing new information. We draw on these insights but add a crucial step: the existence of such related knowledge is necessary for explanation, but not sufficient. This is because the relevant related knowledge must be discovered, and this is not guaranteed in finite time.

Consider an AI system for medical diagnosis that uses random forest models. Even when both essential knowledge elements exist – the doctor understands probability and decision trees, while the AI engineer can articulate the algorithm's logic – discovering the relevant connections within the time constraints of a clinical consultation remains challenging. The engineer might not realize that beginning with ensemble methods (which connects to the doctor's familiarity with seeking second opinions) would be more effective than starting with gradient descent mathematics. In this case, the shared knowledge exists but remains undiscovered within the finite time available for explanation.

The "inmates running the asylum" problem in AI explanation [22] provides another illustration. AI experts often struggle to craft effective explanations for lay users because their deep technical knowledge obscures what others do and do not know. For instance, when explaining a loan denial, an expert might focus on feature importance scores and gradient descent parameters - technically accurate but meaningless to users without machine learning backgrounds. The expert possesses knowledge that could be related to the user's understanding of credit decisions (e.g., drawing parallels between feature importance and traditional credit factors), but discovering these connections within a brief interaction often proves elusive.

These examples highlight a crucial distinction: while prior work emphasizes the necessity of shared knowledge for successful explanation [36], [37], [8], we argue that such overlap, while necessary, is not sufficient. The challenge lies not merely in possessing relevant knowledge, but in discovering the right connections between knowledge elements within finite time constraints. This helps explain why explanations by human experts of AI outputs, or even by the AI itself, may fail; and for humans to use the results of AI, they may simply have to trust it.

We explore this idea using a formal model in the next section.

## 3 MODEL

We model explanation as a search process through networks of knowledge elements. For an explanation to succeed, the explainer must find elements of knowledge (i.e. concepts, topics, beliefs) they share with the explainee and then use these common elements to build a bridge to the new knowledge being conveyed. Just as a teacher might explain a new concept by connecting it to something a student already understands, successful explanation requires discovering points of overlap between the explainer's and explainee's knowledge structures, and building a bridge between this discovered overlap and what is to be explained. Following standard approaches in cognitive science, we formalize this process using network representations, where knowledge elements – concepts, topics, beliefs- are nodes and their relationships are links [37]. For example, in physics, concepts like electric field, electric force, magnetic field or voltage form nodes that are connected through various relationships, from simple associations, explanatory links to more complex experimental procedures and logical deductions [18], [19]. This allows us to analyze precisely how and why explanation attempts may fail even when shared knowledge elements exist.

We consider two actors, an Explainer and Explainee. The knowledge (or beliefs – we do not distinguish these for the purposes of this analysis) of each can be modeled as a graph with nodes as concepts and the undirected edges symbolizing mutual compatibility or coherence between two concepts [32]. We discuss later why our core results do not alter by allowing for negative and weighted ties (e.g., [1]).

To illustrate these ideas, consider, for instance, a molecular biologist (Explainer) attempting to explain protein folding to a computer scientist (Explainee). The biologist's knowledge network might include concepts like 'amino acid sequences,' 'hydrophobic interactions,' and 'protein structure,' with edges between them reflecting their coherent relationships. The computer scientist's network might contain concepts like 'optimization algorithms,' 'energy minimization,' and 'computational complexity,' also with edges reflecting their logical connections. Some concepts might exist in both networks but with different connections – 'energy states' might connect to 'thermodynamics' in the biologist's network but to 'algorithmic efficiency' in the computer scientist's network. For the biologist to successfully explain protein folding, they must find concepts that exist in both networks (like 'optimization' or 'energy states') and use these as bridges to introduce new concepts from their own network.

Formally, the Explainer and Explainee each have a network of knowledge elements represented by networks $R$ (explaineR) and $E$ (explaineE) respectively [37] with their sizes denoted as $N_R$ and $N_E$. These graphs are finite because expanding their size (i.e. adding more nodes) is costly [19], [33]. Therefore, it must also be true that if explanation has succeeded, the benefit to the explainee of adding a node must exceed the cost. Finally, these graphs are initially private – they are mutually unobservable, except through the explanation process itself (which we describe below).

We denote the concept that requires explanation – the target node – by $R_0$ in the Explainer's network $R$. $R$ and $E$ share a common set $K$ with $N_R$ nodes. Note that the set $K$ being non-empty implies that knowledge overlap exists between explainer and explainee [8].

When $R_0$ is in $K$, there is no need for explanation, so we do not consider this case any further. When $R_0$ is not in $K$, providing an explanation of $R_0$ to the Explainee is equivalent to i) finding a node $Y$ in the Explainer's graph $R$ that also exists in the Explainee's graph $E$, (i.e. belongs to $K$); and ii) finding a path in the Explainer's graph $R$ connecting $Y$ to $R_0$.

This formalization captures a fundamental insight about explanation: new knowledge becomes acceptable when it can be connected to existing knowledge. Cognitive psychologists have shown that learners evaluate new information primarily through its coherence with their existing understanding [21]. For instance, Vosniadou and Brewer [39] demonstrate how students accept new scientific concepts by linking them to familiar everyday experiences. Building on this research, we propose that explanation succeeds when the explainer can demonstrate how new knowledge (represented by nodes in their network) connects to knowledge the explainee already possesses (represented by shared nodes).

In formal terms, we assume that once a common node is identified, the explainer can transfer links from this shared node to new nodes in the explainee's network, effectively expanding the latter's knowledge network. This 'bridge-building' process represents how explanations work: by showing how unfamiliar concepts connect to familiar ones in logically coherent ways. Further, we assume that Explainee always accepts a path (however long) from a node in $K$ discovered by the Explainer to the target node (i.e. structure of $E$ does not matter for our analysis). While these are simplifying assumptions – in reality, people may have incentives to resist new knowledge even when logical connections exist, or resist accepting long and complicated explanations even – it allows us to isolate the search challenge in explanation from other such potential barriers.

Thus, to explain the node $R_0$, the objective of the Explainer is to find one explanatory node in $R$ that is connected to $R_0$ through a path (i.e. a chain of nodes). Therefore, we can model an explanation process as a search problem on the Explainer graph $R$. The search process occurs sequentially: at each discrete step, denoted by $t = 1, ..., N_R - 1$, the Explainer selects one node and checks whether it is explanatory (belongs to $K$) or not. If not, the Explainer selects another node connected by a path to a previously visited one (backtracking to the most recent node with unvisited neighbors when no further moves are possible). The search process is thus always local – the Explainer can only move to nodes that connect through a single positive link to any previously examined nodes. This captures how human reasoning typically proceeds through chains of related concepts rather than random jumps between unrelated ideas [37].

We assume that if the Explainer finds an explanatory node, the Explainer and Explainee will receive a one-time benefit $B$. When engaging in the explanation process, at each step $t$, the Explainer and Explainee incur a cost $c(t)$, which is a function of the number of steps $t$ [12]. The incentives of both parties are therefore fully aligned. Further, there is no "tacit node" (i.e. every node in $R$ can be shown by the Explainer as long as it is connected through a positive path to the target node $R_0$), and there is no communication difficulty for Explainee to perceive the node shown by Explainer.

While either party can terminate the explanation process when some stopping rules are triggered, in AI-human interactions where the AI is typically the Explainer and the human is the Explainee, the termination decision is often made by the Explainee. We further assume that the size of the knowledge graph of the Explainer $N_R$ is known by the Explainee. This assumption is plausible for many domain-specific and contemporary AI systems where the system's knowledge scope is well-documented, enabling users to correctly estimate $N_R$. Therefore, both parties have the same knowledge of $N_R$ and face similar uncertainty about the size and location of the overlap set $K$.

## 4 ANALYSIS AND RESULTS

We consider the case when the Explainer's knowledge graph $R$ is a complete graph – all nodes are linked to all other nodes. This rules out path dependence in the explanation process – despite local search, any node is accessible from any other at any point in time in a complete graph.

### 4.1 Expected Time to Achieve Explanation

We first analyze how long the search process may take to achieve explanation when the overlapping set $K$ exists. After the Explainer has shown and failed to explain the target node at $t=0$, we are left with $N_R - 1$ nodes. Let $T$ be the random variable denoting the number of time from $t=1$ until successful explanation. When $R$ is a complete graph, $T$ is equal to the number of time to first success in drawing one node in the overlapping set $K$ without replacement from a finite set $R$. The probability of finding a node in $K$ on the $t^{th}$ draw, given $t-1$ unsuccessful draws is:

$$P(T = t) = \frac{N_K}{(N_R - 1) - (t - 1)} \quad (1)$$

Therefore, distribution of $T$ follows a negative hypergeometric distribution with the expected time to achieve explanation is:

$$E(T) = \frac{N_R}{N_K + 1} \quad (2)$$

For a given $N_R$, this expectation $E(T)$ is a strictly decreasing, convex reciprocal function of $N_K + 1$. Therefore, once $N_K$ passes a threshold, $E(T)$ drops significantly. This suggests that below a critical mass of shared knowledge, the explanation process is time-consuming and thus difficult to achieve. However, once sufficient knowledge overlap exists, the expected time to successful explanation decreases dramatically. This result implies a "knowledge accumulation advantage": agents who achieve the threshold level of shared knowledge can more easily learn and integrate new knowledge, while those below the threshold struggle to achieve explanation.

### 4.2 Early Termination of Explanation

We denote the prior belief of the agent who decides to continue the explanation process (either the Explainer or the Explainee) on the size of the set $K$ (number of overlapping nodes) by $\{i: p_{i0}\}$: from $N_R$ nodes, $i$ nodes can be in the explanatory set $K$ with probability $p_{i0}$. At time $t = 1$, the expected size of the common set K is $\mu_{K1} = \sum_{i=0}^{m} p_{i0}|i|$ and the variance of the size is $V_{K1} = V(i)_1 = \sum_{i=0}^{m} p_{i0}(i - \mu_{K1})^2$.

The decision to initiate the explanation process hinges on the expected benefit at time t=1:

$$E(B_1) = B \frac{\mu_{K1}}{N_R - 1} \quad (1)$$

Assume that the expected benefits are large enough in period 1 to initiate explanation. If an explanation is not achieved, the agent updates the belief about $N_K$ as:

$$p_{i1} = \frac{p_{i0}(N_R - 1 - i)/(N_R - 1)}{1 - \mu_{K1}/((N_R - 1))} = \frac{p_{i0}(N_R - 1 - i)}{N_R - 1 - \mu_{K1}} \quad (2)$$

The expected size of the explanatory set $K$ at time $t=2$ becomes:

$$\mu_{K2} = \mu_{K1} - \frac{V_{K1}}{N_R - 1 - \mu_{K1}} \quad (3)$$

and the expected benefit to continue the explanation process is:

$$E(B_2) = B \frac{\mu_{K2}}{N_R - 2} \quad (4)$$

This calculation can extend to any time $t < N_R$, after revealing *t-1* nodes without finding an explanatory node, with the expected size of K as:

$$\mu_{Kt} = \mu_{K(t-1)} - \frac{V_{K_{t-1}}}{N_R - (t-1) - \mu_{K(t-1)}} \quad (5)$$

and the expected benefit of continuing the explanation process is:

$$E(B_t) = B \frac{\mu_{Kt}}{N_R - t} \quad (6)$$

An implication of (6) is that when explanation begins (i.e. $t$ is $\ll N_R$), the expected benefits of continuing decrease in $N_R$, the size of $R$. This suggests that more knowledgeable agent may see lower benefits to initiating explanation compared to less knowledgeable ones, for the same size of the overlap set.

To examine the behavior of this expected benefit over time, first, we consider the case when the prior belief is uniformly distributed over the support set $\{0, 1, ..., N_R - 1\}$. In this case, the prior mean is $\mu_{K1} = \frac{N_R-1}{2}$, and the prior variance is $V_{K1} = \frac{(N_R-1)(N_R+1)}{12}$. The expected benefit at time *t=1* is $E(B_1) = B/2$.

If the Explainer initiates the explanation process but does not achieve an explanation after the first attempt, the updated mean becomes $\mu_{K2} = \frac{(N_R-1)(N_R-5)}{3(N_R-3)}$, and the expected benefit at time *t=2* is:

$$E(B_2) = \frac{B(N_R - 1)(N_R - 5)}{3(N_R - 3)(N_R - 2)} < E(B_1) \; for \; \forall N_R \quad (7)$$

This result suggests that $E(B_t)$ decreases as *t* increases. To further validate this observation, we examine the behavior of $E(B_t)$ for different values of $N_R$. The simulation results (Figure 1) confirm that for a uniform prior belief, $E(B_t)$ decreases over time.

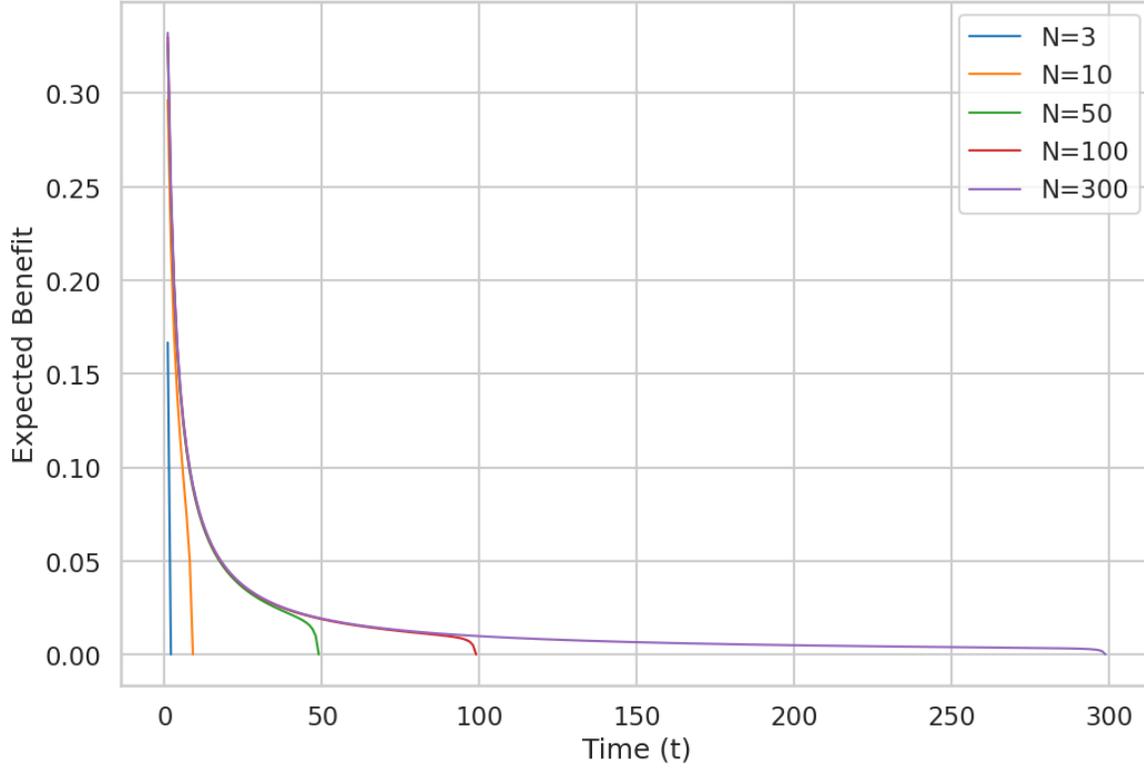

Figure 1. The Dynamics of Expected Benefit Over Time for
Different Sizes of the Explainer' Knowledge Graph under a Uniform Prior Belief

We extend this analysis to consider the general case of any distribution. From equations (1), (3) and (4), we have:

$$E(B_2) > E(B_1) \Leftrightarrow \mu_{K1}(N_R - 1 - \mu_{K1}) > V_{K1}(N_R - 1) \Leftrightarrow 1 - \frac{\mu_{K1}}{N_R - 1} > \frac{V_{K1}}{\mu_{K1}} \quad (8)$$

This condition suggests that when the prior variance $V_{K1}$ is relatively small compared with $\mu_{K1}$ $\left(V_{K1}/\mu_{K1} < 1\right)$, $E(B_2)$ can be higher than $E(B_1)$. Yet, as the prior variance $V_{K1}$ increases beyond a certain value, $E(B_2)$ will become lower than $E(B_1)$. The intuition behind this result is that when the Explainer's initial belief about the size of the common set has a small variance relative to the mean, they are confident in their estimate. When the first attempt to explain is unsuccessful, the Explainer updates their beliefs, but due to this high initial confidence, the failure does not provide enough evidence to shift the belief distribution substantially toward 0. Consequently, the posterior distribution may still place significant probability mass around the prior mean and with a reduction in the denominator, this leads to an increase in the expected benefit.

Conversely, when the variance is large relative to the mean $\left(V_{K1}/\mu_{K1} \geq 1\right)$, the Explainer is less confident about their initial belief about the extent of overlap. An unsuccessful explanation attempt triggers a more substantial downward update in the Explainer's belief about the size of the common set, pushing the distribution further toward 0. The large uncertainty in the initial belief amplifies the impact of the unsuccessful explanation attempt, leading to a decrease in the expected benefit.

However, even when the prior variance is small, $E(B_t)$ does not increase indefinitely. Figure 2 illustrates the temporal dynamics of the expected benefit function for different levels of prior variance relative to the prior mean. The simulation uses a truncated normal distribution over $i = 0 - m$, with knowledge graph size $N_R = 300$ and prior mean fixed at 10. The graphs clearly show that when the variance-to-mean ratio is small (0.5), the expected benefit initially increases before eventually declining. In contrast, with higher variance-to-mean ratios (1.0, 2.0, 5.0), the expected benefit is decreasing in $t$.

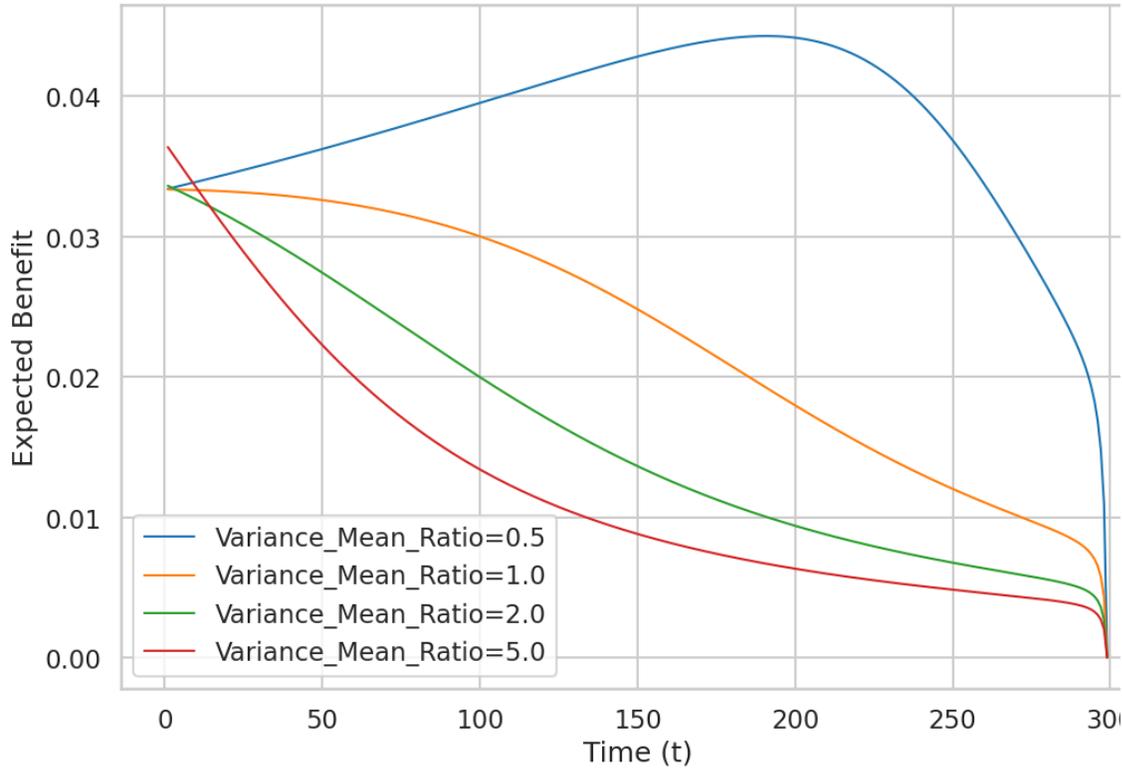

Figure 2. The Dynamics of Expected Benefit Over Time for
Different Levels of Prior Variance

This analysis thus yields two further conclusions. First, the decrease in $E(B_t)$ implies that, at any point in time, if $E(B_t) > c(t)$ is not met, it will be optimal to stop, even though the set $K$ is non-empty. Explanation may fail despite the existence of knowledge overlaps (and by construction in our model – the absence of incentive conflicts, communication problems or tacit knowledge).

Second, the optimal stopping time in the case of greater confidence (i.e. lower prior variance) will be later than in the case of lower confidence (high prior variance), suggesting that having a high confidence in the belief about the size of the overlapping set $K$ can help sustain the explanation process until explainability is achieved.

### 4.3 Summary

Our analysis highlights that explanation may fail even in the presence of substantial knowledge overlap—despite the absence of incentive misalignment, communication barriers, or tacit knowledge constraints in our model. This finding challenges the conventional assumption that shared knowledge necessarily facilitates explanation. Moreover, our results suggest that more knowledgeable Explainers may derive lower expected benefits from initiating explanation compared to their less knowledgeable counterparts, even when the extent of knowledge overlap is identical. However, confidence in one's assessment of the size of the overlap set can play a crucial role in sustaining the explanation process, increasing the likelihood of achieving explainability.

Notably, this confidence can be understood in epistemic terms: a belief that common ground exists even if it is not immediately apparent. Such confidence appears to be instrumental in overcoming early-stage barriers to explanation, allowing the process to continue even in the face of initial difficulty or ambiguity. In this sense, explanation is not merely a function of objective knowledge alignment but also of the Explainer's willingness to persist under uncertainty—a willingness that, paradoxically, may be lower for the more knowledgeable Explainers.

## 5 DISCUSSION AND CONCLUSION

### 5.1 Key Findings

To model explanation - a process that is central to the transfer of knowledge between individuals and the integration of this knowledge – we draw on a well-established formalism in which knowledge elements possessed by an individual can be represented as a network of concepts (e.g., [37], [1]). Explanation, we argued, can be represented as a search process of discovering a node that is common in the networks of the Explainer and Explainee, which has a path to the node the Explainer wants to explain (target node).

Our formal analysis reveals why trust may be necessary precisely when explanation is impossible, even under theoretically ideal conditions. The analysis demonstrates that explanation can fail due to time constraints, even when knowledge overlap exists, positive paths exist between the target concept and shared knowledge, actors are fully motivated and honest, communication is perfect, and all knowledge is explicit. These ideal conditions are in fact approximated quite well in AI-human communication: machines have no incentive to withhold information, can communicate without noise, and can articulate tacit knowledge. Yet explanation remains problematic [30], [31]. This reflects a distinct aspect of bounds on rationality – not processing limits per se but limited time [34]. Further, explainers who are optimistic and confident about the extent of overlap in knowledge are more likely to achieve explanation given overlap, but more knowledgeable Explainers may be less likely to begin explanation at all.

These results have important implications for human-AI interaction. When an AI system produces outputs that require explanation, the search process through their knowledge networks to find appropriate bridging concepts becomes increasingly difficult as their expertise ($N_R$) grows larger, and the expected benefits of continuing the search $E(B_t)$ decline over time unless they have high confidence (low $V_{K1}$) in their estimate of shared knowledge. This creates a fundamental tension: the very depth of expertise that makes an AI system (or indeed human expert) valuable also makes its decisions harder to explain. Users must then make a critical choice between rejecting valuable but unexplainable outputs or accepting them based on trust in the system's competence.

Our analysis suggests this is not a temporary limitation that better explanation techniques will necessarily overcome. Rather, it reflects a fundamental constraint arising from the structure of knowledge networks and the dynamics of search under uncertainty. This provides theoretical grounding for why trust may not just substitute for explanation temporarily, but may be permanently necessary for utilizing complex AI systems - even when they are functioning exactly as intended.

This inevitability of needing to trust AI suggests an important strategic direction for AI development: the need to establish trustworthiness through independent verification mechanisms outside of specific task contexts. Much like how users trust calculators based on their consistent track record of accurate calculations, AI systems might need to build domain-specific reputations for reliability. This implies that AI development efforts should focus not only on improving explanatory capabilities but also on building and maintaining verifiable track records of trustworthiness within specific application domains. Such an approach might favor the development of specialized, domain-specific AI systems over more versatile but less consistently verifiable general-purpose systems.

### 5.2 Extensions

*5.2.1. Partially connected graphs*

Our assumption of a fully connected knowledge graph R for the explainer was intentionally conservative, as it represents the most favorable conditions for successful explanation. By treating R as complete—where every node is connected to every other node —we modeled the Explainer's search process as simple sampling without replacement. At each step t, the Explainer could freely select from the remaining (N_R - t) nodes. However, empirical research in cognitive science suggests that real-world knowledge structures are far more constrained, exhibiting sparse, hierarchical, and complex topologies [18], [19], [6].

When R is not fully connected, the Explainer faces several additional constraints. First, the search becomes locally constrained: at each step t, the Explainer can only examine nodes directly linked to those already visited, preventing them from freely sampling across the network. Second, the search process becomes path-dependent, meaning that early choices shape and constrain future possibilities, increasing the likelihood of inefficient or suboptimal paths. Finally, the Explainer must decide between different search strategies, such as breadth-first search (BFS), which explores nearby nodes, or depth-first search (DFS), which follows specific paths deeper into the network. Without knowing in advance which strategy is most effective for locating nodes in K, this choice introduces further uncertainty and complexity.

These constraints imply that our findings under the assumption of a fully connected R establish a lower bound on the difficulty of explanation. If explanation can fail even in this idealized best-case scenario, it becomes even more challenging when the structure of R reflects the sparsity and complexity observed in real-world knowledge networks.

*5.2.2. Incompatible nodes*

In our modeling, we have focused on compatibility paths between nodes and ignored the possibility of incompatible nodes. However, the formalization we built on also accommodates the coexistence of incompatible beliefs through the following re-interpretation: assume the absence of ties between nodes could arise from two distinct sources. Within any connected component of the network, we interpret zero-strength ties as potentially compatible connections that simply haven't been discovered or made explicit yet. This captures how knowledge elements might be logically coherent with each other, even though the relationships between them haven't been fully mapped out. The search process we modeled represents the discovery of these latent connections.

However, across disconnected components of the network, the absence of ties represents genuine incompatibility between belief clusters. These structural gaps in the network cannot be bridged through any amount of search, reflecting how individuals can maintain contradictory beliefs by compartmentalizing them in disconnected regions of their knowledge structure. This interpretation allows us to explain both how individuals can hold incompatible beliefs and how successful explanation can occur despite the presence of such contradictions.

Consider a climate scientist explaining global warming to a skeptic. The skeptic's knowledge network might contain two disconnected components: one containing scientific concepts and another containing skeptical beliefs. The structural separation between these components represents their fundamental incompatibility. However, explanation can still succeed if the climate scientist finds a path through shared concepts within the scientific component, even while the skeptic maintains contradictory beliefs in the disconnected skeptical component. The key insight is that explanation doesn't require resolving all contradictions across the entire knowledge network - it only requires finding a valid path through shared knowledge within a connected component.

This re-interpretation also allows our model to be compatible with several empirical phenomena: why individuals can learn new concepts in some domains while maintaining contradictory beliefs in others, why explanation attempts might succeed even when actors don't fully resolve all their incompatible beliefs, and how knowledge can grow within domains even while remaining compartmentalized from contradictory belief systems. The structure of the network itself, with its connected and disconnected components, captures both the possibility of discovery within coherent knowledge domains and the persistence of contradictions across them.

A possibly surprising benefit of trust over explanation implied by this conception of incompatibility emerges: explainability-based acceptance of new facts is a force that increases internal connectivity among sub-graphs, whereas acceptance of other's opinions for reasons having to do with trust will produce disjunct sub-graphs. This suggests there is an option value to trust: sole reliance on explanation to accept new knowledge produces internal coherence of what is known but may limit the possibilities of future expansion of what one can know.

### 5.3 Future research directions

Our analysis suggests several promising directions for future research. First, while our model demonstrates how local search constraints impede explanation, a systematic investigation of search strategies (e.g., breadth-first versus depth-first) under varying network topologies could yield valuable insights. Such an analysis would illuminate when explainers should persist with their current approach versus pivoting to alternative strategies, particularly when knowledge networks are only partially connected. This extension could advance our understanding of search processes in both human and artificial intelligence contexts.

Second, our current conceptualization treats knowledge networks as static entities. However, successful explanations likely modify these networks by creating new links within and between components. A dynamic analysis could formalize how learning and explainability co-evolve, with past explanation attempts reshaping network structures in ways that influence future explanations. This relationship between learning and network evolution remains undertheorized in the current literature on organizational learning and knowledge transfer (cf. [2]).

Third, while we focused on dyadic explanation processes, organizational reality often involves multiple explainers and explainees operating simultaneously. Extending our framework to multi-agent settings would allow us to theorize how collective search processes differ from individual ones. This could generate novel insights about collaborative explanation in organizational contexts, particularly regarding how multiple agents might coordinate to bridge knowledge gaps. Such an extension would complement existing work on group versus individual learning (e.g., [5], [15]) by explicitly modeling the search processes involved.

## 5.4 Conclusion

The growing deployment of sophisticated AI systems has intensified concerns about their lack of transparency and explainability. Our analysis reveals that the limits to explanation may be more fundamental than previously recognized, existing even under theoretically ideal conditions of complete knowledge, perfect communication, and aligned incentives. By modeling explanation as a search process through knowledge networks, we show why more knowledgeable explainers may face greater difficulty in explaining their insights, and why trust may be permanently necessary rather than just temporarily useful in human-AI interaction.

## ACKNOWLEDGEMENTS

We thank Vivianna Fang He, Marco Minervini, Sanghyun Park, Tianyu He, Nety Wu and participants in the Organization and Algorithm Meeting for valuable feedback on earlier versions of this paper. The simulation code in the appendix was developed with assistance from Claude 3.5 Sonnet and GPT 4o1, which were used to implement basic computational functions. All results were independently verified for correctness.